\pdfoutput=1

\documentclass[11pt]{article}

\usepackage{naacl2021}

\usepackage{times}
\usepackage{latexsym}
\usepackage[T1]{fontenc}
\usepackage[utf8]{inputenc}
\usepackage{microtype}

\usepackage{cite}
\usepackage{amsmath,amssymb,amsfonts}
\usepackage{algorithmic}
\usepackage{graphicx}
\usepackage{textcomp}
\usepackage{xcolor}
\usepackage{url}
\def\BibTeX{{\rm B\kern-.05em{\sc i\kern-.025em b}\kern-.08em
    T\kern-.1667em\lower.7ex\hbox{E}\kern-.125emX}}
\begin{document}

\title{Human Sentence Processing: Recurrence or Attention?}

\author{Danny Merkx \\
  Radboud University\\ Nijmegen, The Netherlands \\
  {\tt d.merkx@let.ru.nl} \\\And
  Stefan L. Frank \\
  Radboud University\\ Nijmegen, The Netherlands \\
  {\tt s.frank@let.ru.nl} \\}

\maketitle

\begin{abstract}
Recurrent neural networks (RNNs) have long been an architecture of interest for computational models of human sentence processing. The recently introduced Transformer architecture outperforms RNNs on many natural language processing tasks but little is known about its ability to model human language processing.
We compare Transformer- and RNN-based language models' ability to account for measures of human reading effort. 
Our analysis shows Transformers to outperform RNNs in explaining self-paced reading times and neural activity during reading English sentences, challenging the widely held idea that human sentence processing involves recurrent and immediate processing and provides evidence for cue-based retrieval.
\end{abstract}

\section{Introduction}

Recurrent Neural Networks (RNNs) are widely used in psycholinguistics and Natural Language Processing (NLP). Psycholinguists have looked to RNNs as an architecture for modelling human sentence processing (for a recent review, see \citealp{Frank2017}). RNNs have been used to account for the time it takes humans to read the words of a text \citep{Monsalve2012,Goodkind2018} and the size of the N400 event-related brain potential as measured by electroencephalography (EEG) during reading \citep{Frank2015,Rabovsky2016,Brouwer2017,schwartz2019}.

Simple Recurrent Networks (SRNs; \citealp{Elman1990}) have difficulties capturing long-term patterns. Alternative RNN architectures have been proposed that address this issue by adding gating mechanisms that control the flow of information over time; allowing the networks to weigh old and new inputs and memorise or forget information when appropriate. The best known of these are the Long Short-Term Memory (LSTM; \citealp{hochreiter1997long}) and Gated Recurrent Unit (GRU; \citealp{Cho2014}) models. 

In essence, all RNN types process sequential information by recurrence: Each new input is processed and combined with the current hidden state. While gated RNNs achieved state-of-the-art results on NLP tasks such as translation, caption generation and speech recognition \citep{Bahdanau2014,Xu2015,Zeyer2017,Michel2019}, a recent study comparing SRN, GRU and LSTM models' ability to predict human reading times and N400 amplitudes found no significant differences \citep{Aurnhammer2017}.

Unlike the LSTM and GRU, the recently introduced Transformer architecture is not simply an improved type of RNN because it does not use recurrence at all. A Transformer cell as originally proposed \citep{Vaswani2016} consists of self-attention layers \citep{Luong2015} followed by a linear feed forward layer. In contrast to recurrent processing, self-attention layers are allowed to `attend' to parts of previous input directly.

Although the Transformer has 
achieved state-of-the art results on several NLP tasks \citep{Devlin2018,Hayashi2019,Kirata2019}, 
not much is known about how it fares as a model of human sentence processing. The success of RNNs in explaining behavioural and neurophysiological data suggests that something akin to recurrent processing is involved in human sentence processing.
In contrast, the attention operations' direct access to past input, regardless of temporal distance, seems cognitively implausible. 

We compare how accurately the word surprisal estimates by Transformer- and GRU-based language models (LMs) predict human processing effort as measured by self-paced reading, eye tracking and EEG. The same human reading data was used by \citet{Aurnhammer2017} to compare RNN types. We believe the introduction of the Transformer merits a similar comparison because the differences between Transformers and RNNs are more fundamental than among RNN types. All code used for the  training of the neural networks and the analysis is available at \url{https://github.com/DannyMerkx/next_word_prediction}

\section{Background}

\subsection{Human Sentence Processing} 
Why are some words more difficult to process than others? It has long been known that more predictable words are generally read faster and are more likely to be skipped than less predictable words \citep{Ehrlich1981}. Predictability has been formalised as surprisal, which can be derived from LMs. 
Neural network LMs are trained to predict the next word given all previous words in the sequence. After training, the LM can 
assign a probability to a 
word: it 
has an expectation of a word $w$ at position $t$ given the preceding words $w_1, ..., w_{t-1}$. The word's surprisal then equals $-\log P(w_t|w_1, ..., w_{t-1})$.

\citet{Hale2001} and \citet{Levy2008} related surprisal to human word processing effort in sentence comprehension. 
In psycholinguistics, reading times are commonly taken as a measure of word processing difficulty, and the positive correlation between reading time and surprisal has been firmly established \citep{Mitchel2010,Monsalve2012,Smith2013}. 
The N400, a brain potential peaking around 400 ms after stimulus onset and associated with semantic incongruity \citep{Kutas1980}, has been shown to correlate with word surprisal in both EEG and MEG studies \citep{Frank2015,Wehbe2014}.
 
In this paper, we compare RNN and Transformer-based LMs on their ability to predict reading time and N400 amplitude. 
Likewise, \citet{Aurnhammer2017}
compared SRNs, LSTMs and GRUs 
on human reading data from three psycholinguistic experiments. Despite the GRU and LSTM generally outperforming the SRN on NLP tasks, they found no difference in how well the models' surprisal predicted self-paced reading, eye-tracking and EEG data. 

\subsection{Comparing RNN and Transformer} \label{rnn_TF_comp}

\begin{figure}
    \centering
    \includegraphics[width=1\linewidth]{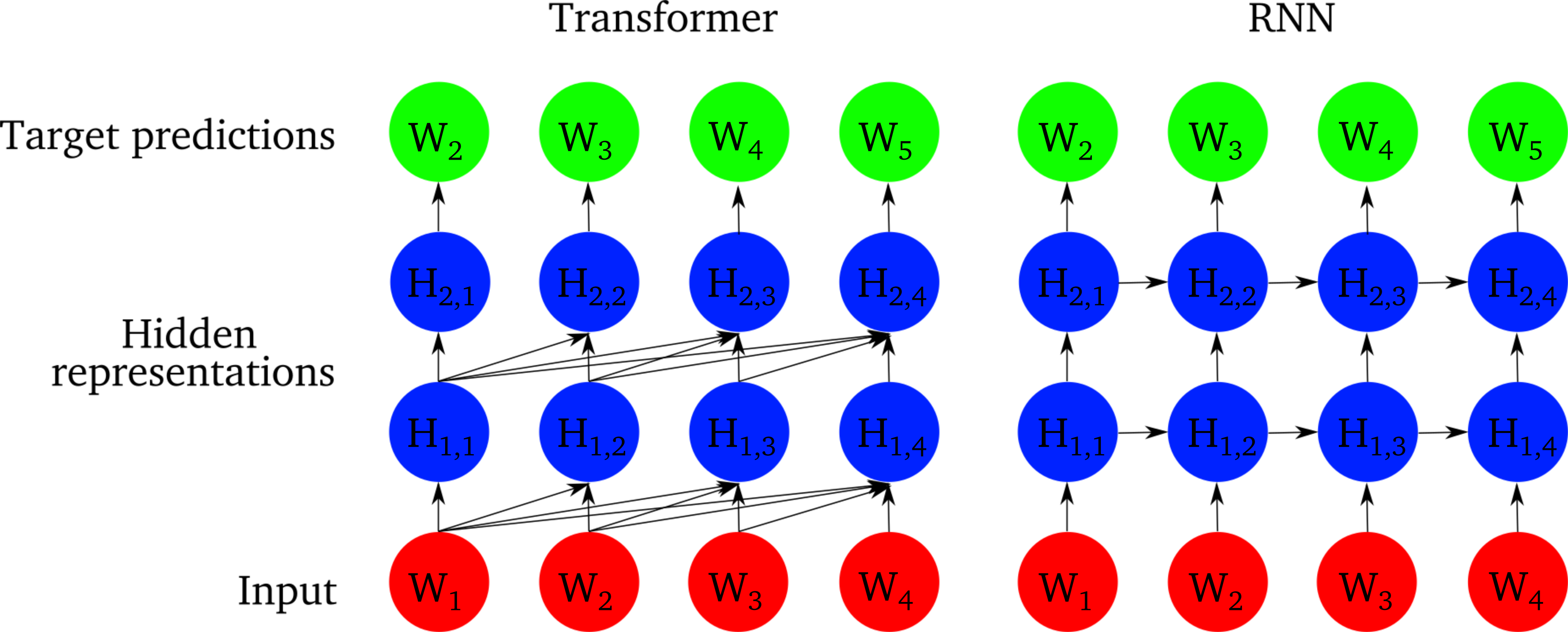}
    \caption{Comparison of sequential information flow through the Transformer and RNN, trained on next-word prediction.} 
    \label{flow}
\end{figure}

According to \citep{Levy2008}, surprisal acts as a `causal bottleneck' in the comprehension process, which implies that predictions of human processing difficulty only depend on the model architecture 
through the estimated word probabilities.
Here we briefly highlight the difference in how RNNs and Transformers process sequential information. The activation flow through the networks is represented in Figure~\ref{flow}.\footnote{Note that the figure only shows how activation is propagated through time and across layers, not how specific architectures compute the hidden states (see \citet{Elman1990,hochreiter1997long,Cho2014,Vaswani2016} for specifics on the SRN, LSTM, GRU and Transformer, respectively).} 

In an RNN, incoming information is immediately processed and represented as a hidden state. The next token in the sequence is again immediately processed and combined with the previous hidden state to form a new hidden state. Across layers, each time-step only sees the corresponding hidden state from the previous layer in addition to the hidden state of the previous time-step, so processing is immediate and incremental. Information from previous time-steps is encoded in the hidden state, which is limited in how much it can encode so decay of previous time-steps is implicit and difficult to avoid. In contrast, the Transformer's attention layer allows each input to directly receive information from all previous time-steps.\footnote{Language modelling is trivial if the model also receives information from future time-steps, as is commonly allowed in Transformers. Our Transformer is thus uni-directional, which is achieved by applying a simple mask to the input.} This basically unlimited memory is a major conceptual difference with RNNs. 
Processing is not incremental over time: Processing of word $w_t$ is not dependent on hidden states $H_1$ through $H_{t-1}$ but on the unprocessed inputs $w_1$ through $w_{t-1}$. Consequently, the Transformer cannot use implicit order information, rather, explicit order information is added to the input. 

However, a uni-directional Transformer can also use implicit order information as long as it has multiple layers. Consider $H_{1,3}$ in the first layer which is based on $w_1, w_2$ and $w_3$. Hidden state $H_{1,3}$ does not depend on the order of the previous inputs (any order will result in the same hidden state). However, $H_{2,3}$ depends on $H_{1,1}, H_{1,2}$ and $H_{1,3}$. If the order of the inputs $w_1,w_2,w_3$ is different, $H_{1,3}$ will be the same hidden state but $H_{1,1}$ and $H_{1,2}$ will not, resulting in a different hidden state at $H_{2,3}$. 

Unlike Transformers, RNNs are inherently sequential, making 
them seemingly more plausible as a cognitive model. \citet{Christiansen2015} argue for a `now-or-never' bottleneck in language processing; incoming inputs need to be rapidly recoded and passed on for further processing to prevent interference from the rapidly incoming stream of new material. In line with this theory, \citet{Futrell2020} proposed a model of lossy-context surprisal based on a lossy representation of memory. Recurrent processing, where input is forgotten as soon as it is processed and only available for subsequent processing through a bounded-size hidden state, is more compatible with these theories than the Transformer's attention operation. 

\section{Methods}

We train LMs with Transformer and GRU architectures and compare how well their surprisal explains human behavioural and neural data. Although a state-of-the-art pre-trained model 
can achieve higher LM quality, we opt to train our own models for several reasons. Firstly, the predictive power of surprisal increases with language model quality \citep{Goodkind2018}, so to separate the effects of LM quality from those of the architectural differences, the architectures must be compared at equal LM capability. We also need to make sure both models have seen the same sentences. Training our own models gives us control over training material, hyper-parameters and LM quality to make a fair comparison.


Perhaps most importantly, we test our models on previously collected human sentence processing data. Most popular large-scale pre-trained models use efficient byte pair encodings as input rather than raw word tokens. 
This is a useful technique for creating the best possible LM, but also a crucial mismatch with how our test material was presented to the human subjects. It is not possible to directly compare the surprisal generated on BPEs to whole-word measures such as gaze durations and reading times. 

\subsection{Language Model Architectures}


We first trained a GRU model using the same architecture as \citet{Aurnhammer2017}: an embedding layer with 400 dimensions per word, a 500-unit GRU layer, followed by a 400-unit linear layer with a tanh activation function, and finally an output layer with log-softmax activation function. All LMs used in this experiment use randomly initialised (i.e., not pre-trained) embedding layers.

We implement the Transformer in PyTorch following \citet{Vaswani2016}. To minimise the differences between the LMs, we picked parameters for the Transformer such that the total number of weights is as close as possible to the GRU model. We also make sure the embedding layers for the models share the same initial weights. The Transformer model has an embedding layer with 400 dimensions per word, followed by a single Transformer layer with 8 attention heads and a fully connected layer with 1024 units, and finally an output layer with log-softmax activation function. 
The total number of parameters for our single-layer GRU and Transformer models are 9,673,137 and 9,581,961 respectively.

We also train two-layer GRU and Transformer models. Neural networks tend to increase in expressiveness with depth 
\citep{abnar2019,giulianelli2018} and a second layer allows the Transformer to use implicit order information, as explained above. While results (see Section \ref{gamcomp}) showed that the two-layer Transformer outperformed the single-layer Transformer in explaining the human reading data, the Transformer did not further benefit from an increase to four layers so we include only the single and two layer models. We did not see a performance increase in the two-layer GRU over the the single-layer GRU and therefore did not try to further increase its layer depth. 

\subsection{Language Model Training}
We train our LMs on Section 1 of the English Corpora from the Web (ENCOW 2014; \citealp{Schafer2015}), consisting of sentences randomly selected from the web. We first exclude word tokens containing numerical values or punctuation other than hyphens and apostrophes, and treat common contractions such as `don't' as a single token. Following \citet{Aurnhammer2017} we then select the 10,000 most frequent word types from ENCOW. 134 word types from the test data (see Section \ref{reading_data}) that were not covered by these most frequent words are added for a final vocabulary of 10,134 words. We select the sentences from ENCOW that consist only of words in the vocabulary and limit the sentence length to 39 tokens (the longest sentence in the test data). Our training data contains 
5.9M sentences with a total of 85M tokens. 


The LMs are trained on a standard next-word prediction task with cross-entropy loss. In the Transformer, we apply a mask to the upper diagonal of the attention matrix such that each position can only attend to itself and previous positions.
To account for random effects of weight initialisation and data presentation order we train eight LMs of each type and share the random seeds between LM types so each random presentation order and embedding layer initialisation is present in both LM types. The LMs were trained for two epochs using stochastic gradient descent with a momentum of 0.9. Initial learning rates (0.02 for the GRU and 0.005 for the Transformer) were chosen such that the language modelling performance of the GRU and Transformer models are as similar as possible. 
The learning rate was halved after $\frac{1}{3}, \frac{2}{3}$, and all sentences during the first epoch and then kept constant over the second epoch. LMs were trained on minibatches of ten sentences.

\subsection{Human Reading Data} \label{reading_data}

We use the self paced reading (SPR, 54 participants) and eye-tracking (ET, 35 participants) data from \citet{Frank2013} and the EEG data (24 participants) from \citet{Frank2015}. In these experiments, participants read English sentences from unpublished novels. In the SPR and EEG experiments, the participants were presented sentences one word at a time. In the SPR experiment the reading was self paced 
while in the EEG experiment words had a fixed presentation time. In the ET experiment, participants were shown full sentences while an eye tracking device monitored which word was fixated. The SPR stimuli consist of 361 sentences, with the EEG and ET stimuli being a subset of the 205 shortest SPR stimuli. The experimental measures representing processing effort of a word are reading time for the SPR data (time between key presses), gaze duration for the ET data (time a word is fixated before the first fixation on a different word) and N400 amplitude for the EEG data (average amplitude at the centroparietal electrodes between 300 and 500 ms after word onset; \citealp{Frank2015}). 

We exclude from analysis sentence-initial and -final words, and words directly followed by a comma. From the SPR and ET data we also exclude the word following a comma, and words with a reading time under 50 ms or over 3500 ms. From the EEG data we exclude datapoints that were marked by \citet{Frank2015} as containing artifacts. 
The numbers of data points for SPR, ET, and EEG were 136,727, 33,001, and 32,417, respectively.

%

\subsection{Analysis Procedure}
At 10 different points during training (1K, 3K, 10K, 30K, 100K, 300K, 1M, 3M sentences and after the first and second epoch) we save each LM's parameters and estimate a surprisal value on each word of the 361 test sentences.

\subsubsection{Linear Mixed Effects Regression}

Following \citet{Aurnhammer2017}, we analyse how well each set of surprisal values predicts the human sentence processing data using linear mixed effects regression (LMER) models with the \textit{MixedModels} package in \textsf{Julia} \citep{douglas2019}. 
For each datasets (SPR, ET, and EEG) we fit a baseline LMER model which takes into account several factors known to influence processing effort. The dependent variables for the SPR and ET models are log-transformed reading time and gaze duration, respectively; for the EEG model it is the size of the N400 response. All LMER models include log-transformed word frequency (taken from SUBTLEXus; \citealp{Brysbaert2009}), word length (in characters) and the word's position in the sentence 
as fixed effects. 

Spill-over 
occurs
when processing a word is not yet completed when the next word is read \citep{Rayner}.
To account for spill-over in the SPR and ET data we include the previous word's frequency and length. For the SPR data, we include the previous word's reading time to account for the high correlation between consecutive words' reading times. 
For the EEG data, we include the baseline activity (average amplitude in the 100 ms before word onset). All fixed effects were standardised, and all LMER models include two-way interaction effects between all fixed effects, by-subject and by-item (word token) random intercepts, and by-subject random slopes for all fixed effects. 

After fitting the baseline models, we 
include the surprisal values (for SPR and ET also the previous word's surprisal) as fixed effects, but no new interactions. 
For each LMER model with surprisal, we calculate the log-likelihood ratio with its corresponding baseline model, indicating the decrease in model deviance due to adding the surprisal measures. The more the surprisal values decrease the model deviance, the better they predict the human reading data. We call this log-likelihood ratio the goodness-of-fit between the surprisal and the data. 
Surprisal from the early stages of training often received a negative coefficient, contrary to the expected longer reading times and higher N400 amplitude for higher surprisal.
This could be caused by collinearity, most likely between surprisal and the log-frequency, which was confirmed by their very high correlation ($>.9$) and Variance Inflation Factors ($>15$) \citep{Tomaschek2018}. 
Apparently, the neural networks 
are very sensitive to word frequency before they learn to pick up on more complex relations in the data. 
We indicate affected goodness-of-fit scores by adding a negative sign and excluded these scores from the next stage of analysis. 

\subsubsection{Generalised Additive Modelling}

As said before, it is well known that surprisal values derived from better LMs are a better fit to human reading data \citep{Monsalve2012,Frank2015,Goodkind2018}. 
We use generalised additive modelling (GAM) to assess whether the LMs differ in their ability to explain the human reading data at equal language modelling capability, that is, because of their architectural differences and not due to being a better LM. The log-likelihood ratios obtained in the 
LMER analyses
are a measure of how well each LM explains the human reading data. We use each LM's average log probability 
over the datapoints used in the LMER analyses as a measure of the LM's language modelling capability. Separate GAMs are fit for each of the three datasets, using the \textsf{R} package \textit{mgcv} by \citep{Wood2004}. LM type (single-layer GRU, two-layer GRU, etc.) is used as an unordered factor so that separate smooths are fit for each LM type. Furthermore, we add training repetition (i.e., the random training order and embedding initialisation) as a random smooth effect.

\section{Results}

\subsection{LM Quality and Goodness-of-Fit}

Figure \ref{gam} shows the goodness-of-fit values 
from the LMER models and the smooths fit by the GAMs. 
Overall we see the expected relationship where higher LM quality results in higher goodness of fit. The LM quality increases monotonically during training, meaning the clusters seen in the scatter-plots correspond to the points during training where the network parameters were stored.
The models do seem to reach similar levels of LM quality at the end of training: The average log probability of the best LM (two-layer Transformer) is only 0.17 higher than the worst LM (two-layer GRU).

\subsection{GAM Comparisons} \label{gamcomp}

\begin{figure*}
    \centering
    \includegraphics[width=.9\linewidth]{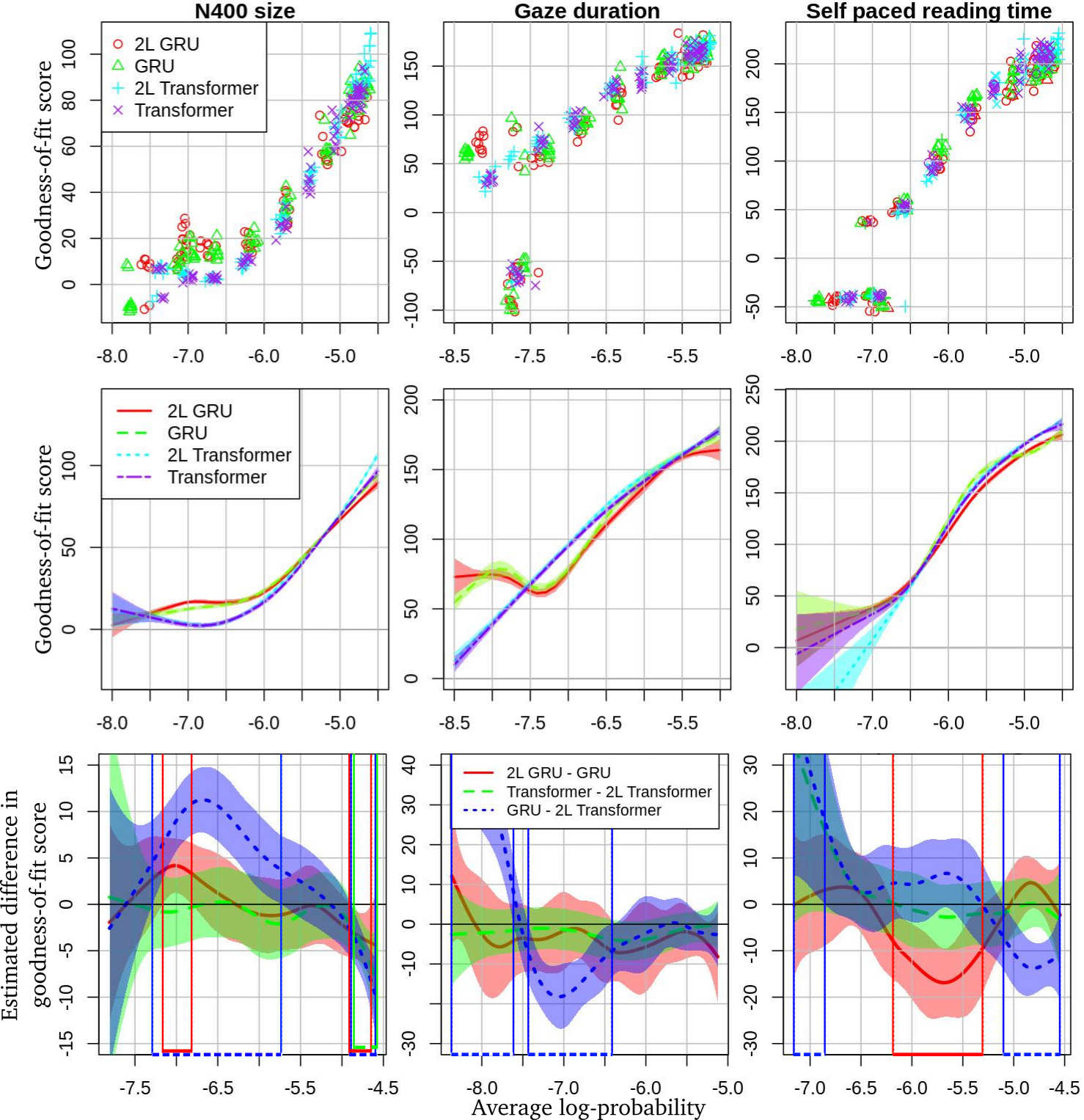}
    \caption{Top row: results of the linear mixed effects regression analysis grouped by LM type. These scatter-plots show the resulting goodness-of-fit values plotted against the average log-probability over the included test data. Negative goodness-of-fit indicates effects in the unexpected direction. Middle row: smooths resulting from the GAMs fitted on the goodness-of-fit data (excluding negative values), with their 95\% confidence intervals.
    Bottom row: estimated differences in goodness-of-fit score. The markings on the x-axis and the vertical lines indicate intervals where zero is not within the 95\% confidence interval. Each curve represents a comparison between two models, with an estimated difference above zero meaning the first model performed better and vice versa for differences below zero.}
    \label{gam}
\end{figure*}

The bottom row of Figure \ref{gam} shows the estimated differences between the GAM curves in the middle row. 
The two-layer GRU does not seem to improve over the single-layer GRU. It outperforms the single-layer GRU only in the early stages of training on the EEG data, with the single-layer GRU outperforming it in the later stages and on the SPR data. The two-layer GRU also reaches lower LM quality on all datasets. For the Transformers we see the opposite, with the two-layer Transformer outperforming the single-layer Transformer on the N400 data at the end of training and never being outperformed by its shallower counterpart. The two-layer Transformer reaches a higher maximum LM quality on all datasets.

For the comparison between architectures, we only compare the best model of each type, i.e., the single-layer GRU and two-layer Transformer. The GRU outperforms the Transformer in the early stages of training (3K-300K sentences) on the N400 data, but the Transformer outperforms the GRU at the end of training on both the SPR and N400 data. 
On the gaze duration data, there are some performance differences with the Transformers and GRUs outperforming each other at different points during training but there are no differences in the later stages of training.

\subsection{Shorter and Longer Sentences in SPR}

\begin{figure*}
    \centering
    \includegraphics[width=.9\linewidth]{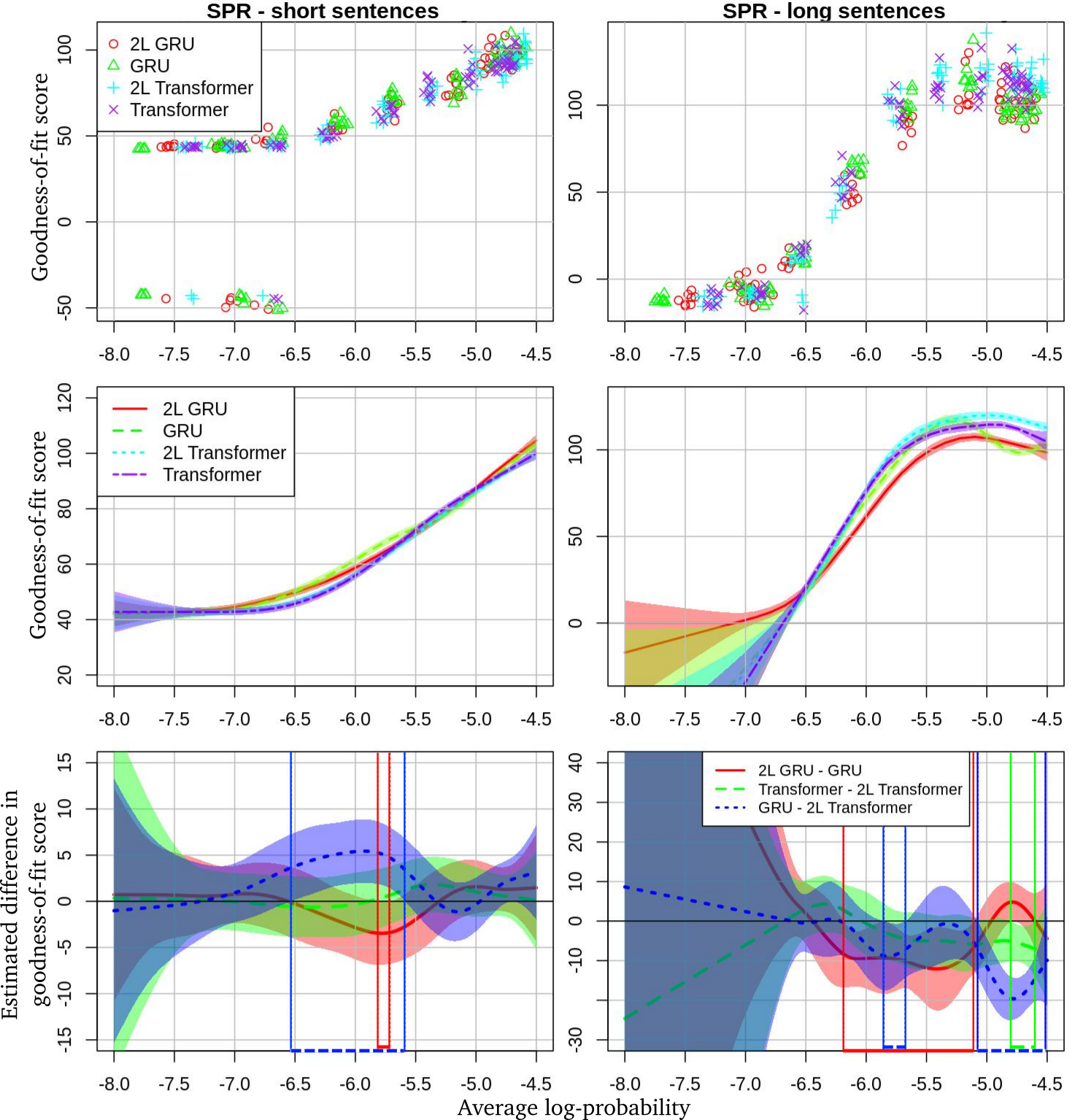}
    \caption{Top row: the results of the linear mixed effects regression analysis on the SPR data, where the data is split by whether the sentences were present in the ET/EEG experiment or not. These scatter-plots show the resulting goodness-of-fit values plotted against the average surprisal over the included test data. Middle row: the smooths resulting from the GAMs fitted on the goodness-of-fit data, with their 95\% confidence intervals. Bottom row: the estimated differences in goodness-of-fit score with intervals where 0 is not withing the 95\% confidence interval marked by vertical lines and markers on the x-axis. Each curve represents a comparison between two models, with an estimated difference above zero meaning the first model performed better and vice versa for differences below zero.}
    \label{gam_app_C}
\end{figure*}

The SPR data contains a subset of sentences longer (in number of characters) than those in the EEG/ET data. As the Transformer has unlimited memory of past inputs, the presence of longer sentences could explain why it outperformed the GRU on the SPR data. We repeated the analysis of the single- and two-layer GRUs and Transformers but only on those sentences from the SPR data that also occurred in the EEG/ET data. On these shorter sentences, there are no notable performance differences between any of the LM architectures (Figure \ref{gam_app_C}). When we test on only those sentences that were not included in the EEG/ET experiments (i.e., the longer sentences), the Transformers outperform the GRUs as they did on the complete SPR dataset. 

\section{Discussion}

We trained several language models based on Transformer and GRU architectures to investigate how well these neural networks account for human reading data. At equal LM quality, the Transformers generally outperform the GRUs. 
It seems that their attention-based computation allows them to better fit the self-paced reading and EEG data. This is an unexpected result, as we considered the Transformer's unlimited memory and access to past inputs implausible given current theories on human language processing.

Notably, the Transformer outperformed the GRU on the two datasets where sentences were presented to participants word by word (SPR and EEG). Neurophysiological evidence suggests that natural (whole sentence) reading places different demands on the reader than word-by-word reading, leading to different encoding and reading strategies \citep{Metzner2015}. Metzner et al. speculate that word-by-word reading places greater demand on working memory, leading to faster retrieval of previously processed material. This seems to be supported by our results; the Transformer has direct access to previous inputs and hidden states and is better at explaining the RT and N400 data from the word-by-word reading experiments. However, when we split the SPR data by sentence length, the results suggest that the Transformers' advantage is mainly due to performing better on the longer SPR sentences. On the other hand, the Transformer did outperform the GRU on the EEG dataset which contains only the shorter subset of sentences. The question remains whether the Transformer's unlimited memory is an advantage on longer sentences only, or if it could also explain why it performs better on data presented word-by-word. 
This question could be resolved with new data gathered in experiments where the same set of stimuli is used in SPR and EEG. Furthermore, future research could do a more specific error analysis to identify on which sentences the Transformer performs better, and perhaps even on which sentences the GRU performs better. Such an analysis may reveal the models are sensitive to certain linguistic properties allowing us to form testable hypotheses.

Surprisingly, adding a GRU layer did not improve performance, and even hurt it on reading time data, despite previous research showing that increasing layer depth in RNNs allows them to capture more complex patterns in linguistic data \citep{abnar2019,giulianelli2018}. The Transformers did show improvement when adding a second layer but did not improve much with four layers. As explained in Section 2, a single-layer Transformer cannot make use of implicit order information in the sequence. 
When adding a single layer to our Transformer, the second layer operates no longer on raw input embeddings but on contextualised hidden states allowing the model to utilise implicit input order information. Further layers increase the complexity of the model but do not make such a fundamental difference in how input is processed. In future work we could investigate how powerful this implicit order information is, and whether multi-layer Transformer LMs no longer require the additional explicit order information.

Our results raise the question how good recurrent models are as models of human sentence processing if they are outperformed by a cognitively implausible model. However, one could also interpret the results in favour of Transformers (and the attention mechanism) being plausible as a cognitive model. While unlimited working memory is certainly implausible, some argue that the capacity of working memory is even smaller than often thought (only 2 or 3 items) and that language processing depends on rapid direct-access retrieval of items from storage \citep{mcelree1998,lewis2006}. Cue based retrieval theory posits that items are rapidly retrieved based on how well they match the cue \citep{parker2017}. This is compatible with the attention mechanism used in Transformers which, simply put, weighs previous inputs based on their similarity to the current input. Cue-based retrieval models due have a recency bias due to decaying activation of memory representations but it is possible to implement a similar mechanism in Transformers \citep{peng2021}.

Interestingly, \citet{lewis2006} claims that serial order information is retrieved too slowly to support sentence comprehension. However, our two-layer Transformer outperforms the single layer Transformer, presumably due to order information implicitly arising as a natural result from the attention operation being performed. The use of serial order information could be compatible with cue-based retrieval models if the order information can naturally arise from the retrieval operations. 

In conclusion, we investigated how the Transformer architecture holds up as a model of human sentence processing compared to the GRU. Our Transformer LMs are better at explaining the EEG and SPR data which contradicts the widely held idea that human sentence processing involves recurrent and immediate processing with lossy retrieval of previous input and provides evidence for cue-based retrieval in sentence processing. 

\section*{Acknowledgements}

The research presented here was funded by the Netherlands Organisation for Scientific Research (NWO) Gravitation Grant 024.001.006 to the Language in Interaction Consortium.

\bibliographystyle{acl_natbib}
\bibliography{mybib}

\end{document}